\pdfoutput=1

\documentclass[11pt]{article}

\usepackage{EMNLP2022}

\usepackage{times}
\usepackage{latexsym}

\usepackage[T1]{fontenc}

\usepackage[utf8]{inputenc}

\usepackage{microtype}

\usepackage{inconsolata}

\usepackage{multirow}
\usepackage{bbding} 
\usepackage{booktabs} 
\usepackage{graphicx} 
\usepackage{arydshln} 
\usepackage{lipsum}  
\usepackage{amsmath}
\usepackage{enumitem} 
\usepackage{bm}
\usepackage{amsfonts}
\usepackage{pifont}

\definecolor{mygray}{gray}{0.6}
\definecolor{ForestGreen}{RGB}{34,139,34}

\newcommand{\ie}{\textit{i}.\textit{e}.}
\newcommand{\eg}{\textit{e}.\textit{g}.}
\usepackage{wrapfig}

\usepackage{xparse}
\NewDocumentCommand{\heng}
{ mO{} }{\textcolor{red}{\textsuperscript{\textit{Heng}}\textsf{\textbf{\small[#1]}}}}
\NewDocumentCommand{\yulei}
{ mO{} }{\textcolor{blue}{\textsuperscript{\textit{Yulei}}\textsf{\textbf{\small[#1]}}}}
\NewDocumentCommand{\hammad}
{ mO{} }{\textcolor{green}{\textsuperscript{\textit{Hammad}}\textsf{\textbf{\small[#1]}}}}
\NewDocumentCommand{\xd}
{ mO{} }{\textcolor{green}{\textsuperscript{\textit{Xudong}}\textsf{\textbf{\small[#1]}}}}

\title{Weakly-Supervised Temporal Article Grounding}

\author{
Long Chen$^\heartsuit$, Yulei Niu$^\heartsuit$, Brian Chen$^\heartsuit$, Xudong Lin$^\heartsuit$, Guangxing Han$^\heartsuit$, \\ \textbf{Christopher Thomas$^\diamondsuit$, Hammad Ayyubi$^\heartsuit$, Heng Ji$^\spadesuit$, and Shih-Fu Chang$^\heartsuit$} \\
$^\heartsuit$Columbia University \;
$^\diamondsuit$Virginia Tech \;
$^\spadesuit$University of Illinois at Urbana-Champaign \\
\texttt{\{cl3695, yn2338, bc2754, xl2798, gh2561, ha2578, sc250\}@columbia.edu} \\
\texttt{chris@cs.vt.edu}, \; \texttt{hengji@illinois.edu}
}

\begin{document}
\maketitle

\begin{abstract}
Given a long untrimmed video and natural language queries, video grounding (VG) aims to temporally localize the semantically-aligned video segments. Almost all existing VG work holds two simple but unrealistic assumptions: 1) \emph{All query sentences can be grounded in the corresponding video}. 2) \emph{All query sentences for the same video are always at the same semantic scale}. Unfortunately, both assumptions make today's VG models fail to work in practice. For example, in real-world multimodal assets (\eg, news articles), most of the sentences in the article can not be grounded in their affiliated videos, and they typically have rich hierarchical relations (at different semantic scales). To this end, we propose a new challenging grounding task: \emph{Weakly-Supervised temporal Article Grounding} (WSAG). Specifically, given an article and a relevant video, WSAG aims to localize all ``groundable'' sentences to the video, and these sentences are possibly at different semantic scales. Accordingly, we collect the first WSAG dataset to facilitate this task: \textbf{YouwikiHow}, which borrows the inherent multi-scale descriptions in wikiHow articles and plentiful YouTube videos. In addition, we propose a simple but effective method \textbf{DualMIL} for WSAG, which consists of a two-level MIL\footnote{MIL: Multiple Instance Learning.} loss and a single-/cross- sentence constraint loss. These training objectives are carefully designed for these relaxed assumptions. Extensive ablations have verified the effectiveness of DualMIL\footnote{Codes: \href{https://github.com/zjuchenlong/WSAG}{https://github.com/zjuchenlong/WSAG}.}.
\end{abstract}

\section{Introduction}
Video Grounding (VG), \ie, localizing video segments that semantically correspond to (coreference relation) query sentences, is one of the fundamental tasks in multimodal understanding. Further, video grounding can serve as an indispensable technique for many downstream applications, such as the text-oriented highlight detection~\cite{lei2021qvhighlights}, video retrieval~\cite{miech2020end} or video question answering~\cite{ye2017video,xiao2022rethinking}.

\begin{figure}[t]
	\centering
	\includegraphics[width=0.99\linewidth]{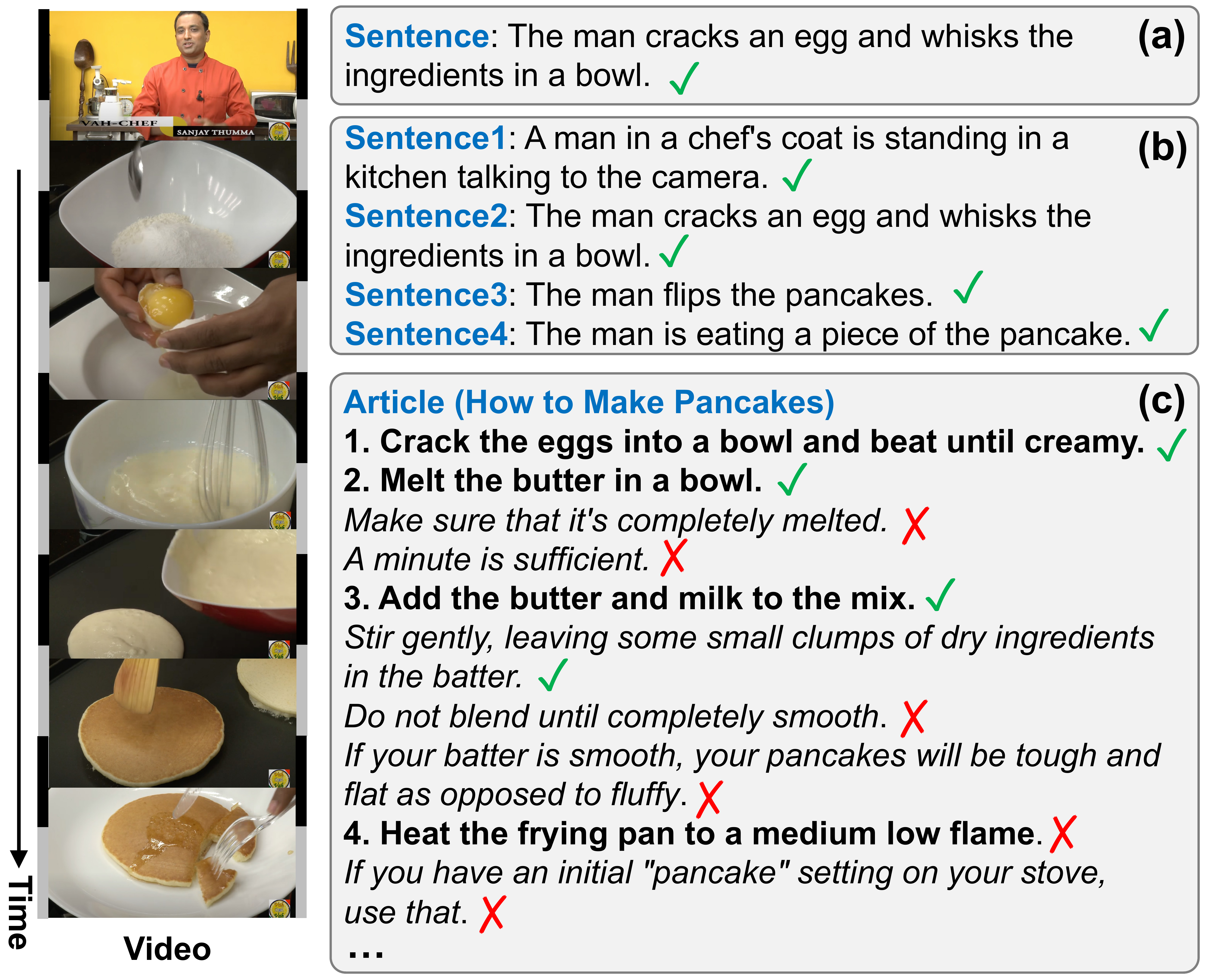}
	\vspace{-2em}
	\caption{(a) Single sentence grounding: The query is a single sentence. (b) Multi-sentence grounding: The queries are multiple sentences. (c) Article grounding: The query is an article, which consists of multiple sentences at different scales (\eg, \href{https://www.wikihow.com/Make-Pancakes}{\emph{How to Make Pancakes}}). \textbf{High-level} and \textit{low-level} sentences are denoted with corresponding formats. \textcolor{ForestGreen}{\ding{51}} and \textcolor{red}{\ding{55}} denote that sentence can or cannot be grounded to the video, respectively.}
	\label{fig:1}
\end{figure}

Early VG efforts mainly focus on single sentence grounding~\cite{gao2017tall,anne2017localizing} (cf. Figure~\ref{fig:1}(a)). Thanks to advanced representation learning and multimodal fusion techniques, single sentence VG has achieved unprecedented progress over the recent years~\cite{cao2021pursuit}. The next step towards general VG is to ground multiple sentences to the same video (cf. Figure~\ref{fig:1}(b)). A straightforward solution for multi-sentence VG is utilizing the single sentence VG model for each sentence individually. Since these query sentences associated with the same video are always semantically related, recent multi-sentence VG methods directly ground all queries simultaneously by considering their temporal order or semantic relations~\cite{bao2021dense,shi2021end}.

Unfortunately, all existing VG attempts hold two simple but unrealistic assumptions: 1) \textbf{\emph{All query sentences can be grounded in the corresponding video}}. Although this assumption is acceptable for the VG task itself, it greatly limits the usage of VG models in real-world multimodal assets. For example in news articles, most of the sentences in an article cannot be grounded in their affiliated videos. 2) \textbf{\emph{All query sentences for the same video are always at the same semantic scale}}. By ``same scale'', we mean that all VG models overlook the hierarchical (or subevent) relations~\cite{aldawsari2019detecting,yao2020weakly} between these query sentences. For example, in Figure~\ref{fig:1}(c), the sentence ``\textit{Stir gently, leaving some small clumps of dry ingredients in the batter}'' ($S_2$) is one of the subevents of ``\textit{Add the butter and milk to the mix}'' ($S_1$), \ie, $S_1$ and $S_2$ are at different semantic scales. Thus, the second assumption makes current VG models fail to perceive the semantic scales, and achieve unsatisfactory performance with multi-scale queries.

\begin{figure}[t]
\centering
  \includegraphics[width=0.98\linewidth]{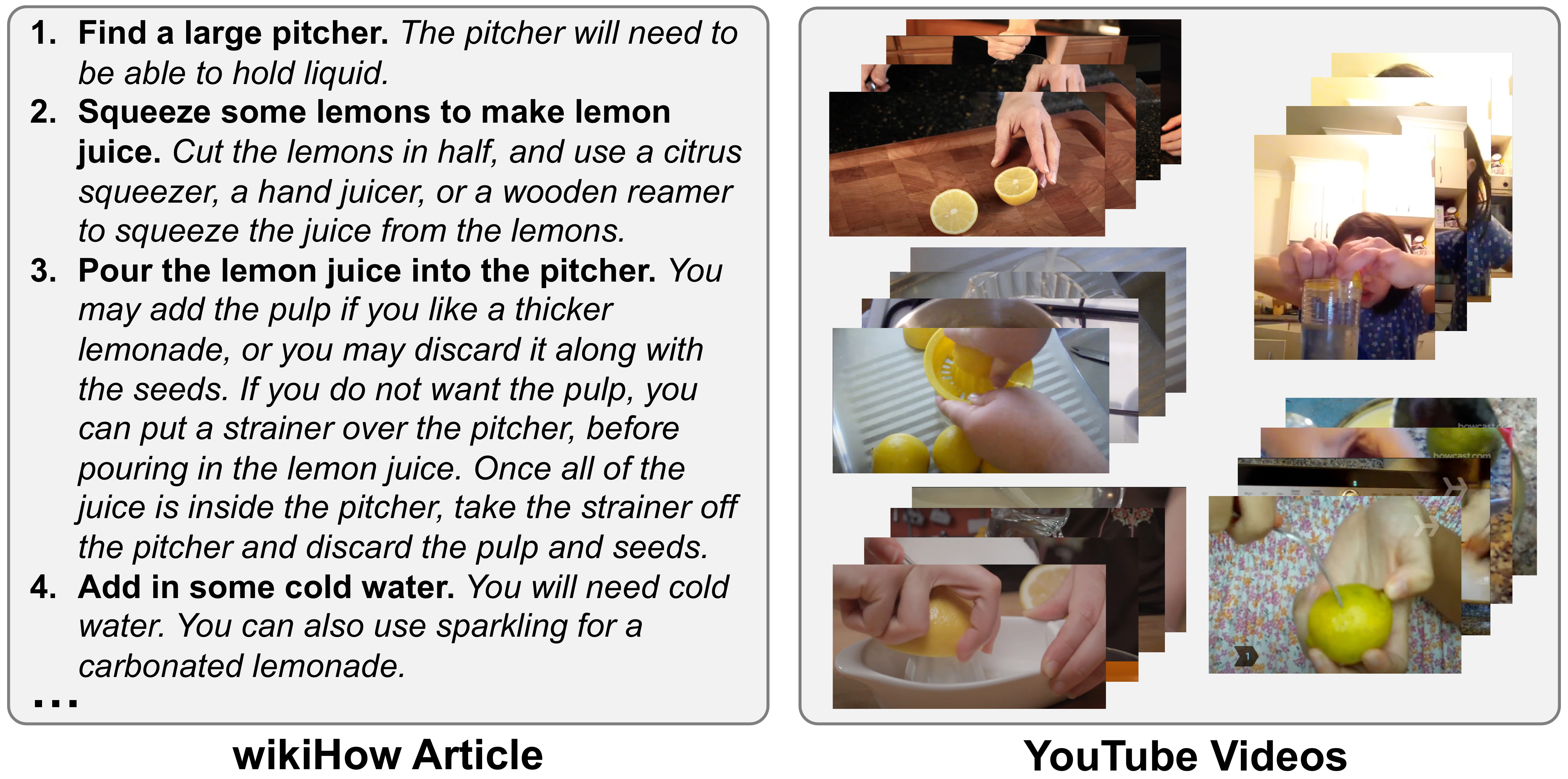}
  \vspace{-0.6em}
  \caption{The only supervision for WSAG is a wikiHow article (\eg, \textit{How to Make Lemonade}) and some corresponding YouTube videos about the same task.}
  \label{fig:2}
\end{figure}

To this end, we propose a more realistic but challenging grounding task: \textbf{Article Grounding} (AG), which relaxes both above-mentioned assumptions. Specifically, given a video and a relevant article (\ie, a sequence of sentences), AG requires the model to localize only ``groundable'' sentences to video segments, and these sentences are possibly at different semantic scales. To further avoid the manual annotations for the large-scale training set, in this paper, we consider a more meaningful setting: weakly-supervised AG (\textbf{WSAG}). As shown in Figure~\ref{fig:2}, the only supervision for WSAG is that the given video and article are about the same task\footnote{A task means the same topic with clear and specific steps.}.

Since there is no prior work on WSAG, we collect a new dataset, \textbf{YouwikiHow}, to benchmark the research. YouwikiHow is built on top of wikiHow articles and YouTube videos\footnote{\href{https://www.wikihow.com/}{https://www.wikihow.com/} \&  \href{https://www.youtube.com/}{https://www.youtube.com/}.}. In particular, we group a wikiHow article and an arbitrary video about the same \emph{task} as a document-level pair (cf. Figure~\ref{fig:2}). For the training set, we conduct a set of carefully designed operations to control the quality of training samples, \eg, task filtering or sentence simplification. For the test set, we directly borrow the manual step grounding annotations in the existing CrossTask~\cite{zhukov2019cross} dataset and propagate them to wikiHow article sentences.

In addition, we propose a simple but effective Dual loss constraint MIL-based method for WSAG, dubbed \textbf{DualMIL}. Specifically, \textbf{for the first assumption}, we relax the widely-used Multiple Instance Learning (MIL) loss into a two-level MIL loss. By ``two-level'', we mean that we regard \emph{all sentences for each article} (sentence-level) and \emph{all proposals for each sentence} (segment-level) as the ``bag'' at two different levels. Then, we obtain the global video-article matching score by aggregating all matching scores over the two-level bag. This two-level MIL inherently allows some queries that cannot be grounded in the video. Meanwhile, to avoid obtaining many highly-overlapping segments, we propose a single-sentence constraint to suppress the proposals whose neighbor proposals have higher matching scores with the query. \textbf{For the second assumption}, we enhance models' abilities in perceiving different semantic scale queries by considering these hierarchical relations across sentences. In particular, we assume that high-level sentences should be more likely to be grounded than its low-level sentences for highly matched proposals, and propose a cross-sentence constraint loss. We show the effectiveness of DualMIL over state-of-the-art methods through extensive ablations.

In summary, we make three contributions:
\begin{enumerate}[leftmargin=1em]
\vspace{-0.5em}
\itemsep-0.4em

\item To the best of our knowledge, we are the first work to discuss the two unrealistic assumptions: all query sentences are groundable and all query sentences are at the same semantic scale. Meanwhile, we propose a meaningful WSAG task.

\item To benchmark the research, we collect the first WSAG dataset: YouwikiHow.
    
\item We further propose a simple but effective method DualMIL for WSAG, which consists of three different model-agnostic training objectives.
\end{enumerate}

\section{Related Work}

\noindent\textbf{Single Sentence \& Multi-Sentence VG.}
Mainstream solutions for single sentence VG can be coarsely categorized into two groups: 1) \textit{Top-down Methods}~\cite{anne2017localizing,gao2017tall,zhang2019man,zhang2020learning,chen2018temporally,yuan2019semantic,yuan2021closer,wang2020temporally,xiao2021boundary,xiao2021natural,liu2021context,liu2021adaptive,lan2022closer}: They first cut given video into a set of segment proposals with different durations, and then calculate matching scores between query and all segment proposals. Their performance heavily relies on predefined rules for proposal settings (\eg, temporal sizes). 2) \textit{Bottom-up Methods}~\cite{yuan2019find,lu2019debug,zeng2020dense,chen2020rethinking,chen2018temporally,zhang2020span}: They directly predict the two temporal boundaries of the target segment by regarding the query as a conditional input. Compared to their top-down counterpart, bottom-up methods always fail to consider the global context between two boundaries (\ie, inside segment). In this paper, we follow the top-down framework and our DualMIL is model-agnostic.

Existing multi-sentence VG work all takes an assumption: the query sentences are ranked by their corresponding segments. This is an unrealistic and artificial setting. In contrast, real-world articles always do not meet this strict requirement, and most of the sentences are not even groundable in affiliated videos. In this paper, we take more realistic assumptions for the multi-sentence VG problem.

\noindent\textbf{Weakly-Supervised VG.}
Since the agreements on the manually annotated target segments tend to be low~\cite{otani2020uncovering}, a surge of efforts aims to solve this challenging task in a weakly-supervised manner, \ie, there are only video-level supervisions at the training stage. Currently, there are two typical frameworks: 1) \textit{MIL-based}~\cite{gao2019wslln,mithun2019weakly,chen2020look,ma2020vlanet,zhang2020regularized,zhang2020counterfactual,tan2021logan}: They first calculate the matching scores between the query sentence and all segment proposals and then aggregate scores of multiple proposals as the score of whole ``bag''. State-of-the-art MIL-based methods usually focus on designing better positive/negative bag selections. 2) \textit{Reconstruction-based}~\cite{duan2018weakly,lin2020weakly}: They utilize the consistency between dual tasks \emph{sentence localization} and \emph{caption generation}, and infer the final grounding results from intermediate attention weights. Among them, the most related work to us is CRM~\cite{huang2021cross}, which considers both multi-sentence and weakly-supervised settings. Compared to CRM, our setting is more challenging: a) Sentences are from different scales; b) Not all sentences can be groundable; and c) Sentence sequences are not consistent with GTs.
 
\noindent\textbf{Multi-Scale VL Benchmarks.} With the development of large-scale annotation tools, hundreds of video-language (VL) datasets are proposed. To the best of our knowledge, three (types of) VL datasets also have considered the multiple semantic scale issue: 1) \textit{TACoS Multi-Level}~\cite{rohrbach2014coherent}: It provides three-level summaries for videos. In contrast, their middle-level sentences are more like extractive summarization (instead of abstractive). Thus, the grounding results for different-scale sentences may be the same. 2) \textit{Movie-related}~\cite{xiong2019graph,huang2020movienet,bain2020condensed}: They always have multiple-level sentences to describe videos, such as overview, storyline, plot, and synopsis. They have two characteristics: a) Numerous sentences are abstract descriptions, \ie, they do not have exact grounding temporal boundaries. b) The high-level summaries are more like highlights or salient events. 3) \textit{COIN}~\cite{tang2019coin}: It defines multi-level predefined steps. Thus, it sacrifices the ability to ground any open-ended queries.

\addtolength{\tabcolsep}{-3pt}
\begin{table*}[t]
\small
    \centering
    \scalebox{0.96}{
        \begin{tabular}{l c c c c c c c}
        \hline
            \multirow{2}{*}{Dataset} &  \multirow{2}{*}{\#Videos} & {Avg Sents} & \multirow{2}{*}{\#Tasks} & Multi-Moment & {Open} & {Support} & {May Not} \\
            & & {per Video} & & {per Query} & {Vocabulary} & {Multi-Scale} & {Groundable} \\
        \hline
            DiDeMo~\cite{anne2017localizing} & 10.6K & 3.9 & --- & --- &  \Checkmark & --- & --- \\
            Charades-STA~\cite{gao2017tall} & 6.7K & 2.4 & --- & --- &  \Checkmark & --- & --- \\
            ANet-Caps~\cite{krishna2017dense} & 15K & 4.8 & --- & --- & \Checkmark & --- & --- \\
            YouCook2~\cite{zhou2018towards} & 2K & 7.7 &  89 & \Checkmark & --- & --- & --- \\
            TVR~\cite{lei2020tvr} & 21.8K & 5.0 & --- & --- &  \Checkmark & --- & --- \\
            QVHighlights~\cite{lei2021qvhighlights} & 10.2K & 1.0 & --- & \Checkmark & \Checkmark & --- & --- \\
        \hline
            CrossTask~\cite{zhukov2019cross} & 4.7K & 7.4$\sim$8.8 & 83 & \Checkmark & --- & --- & --- \\
            COIN~\cite{tang2019coin} & 11.8K & 3.9 & 180 & --- & --- & \Checkmark & --- \\
        \hline
            YouwikiHow (training set) & 47K & 20.8 & 1,398 & \Checkmark & \Checkmark & \Checkmark & \Checkmark \\
        \hline
        \end{tabular}
    }
    \vspace{-0.5em}
    \caption{Comparison between YouwikiHow and other prevalent video grounding or step segmentation benchmarks.}
    \label{tab:dataset_comparison}
\end{table*}
\addtolength{\tabcolsep}{3pt}

\section{Dataset: YouwikiHow}
We built \textbf{YouwikiHow} dataset from wikiHow articles and YouTube videos. As shown in Figure~\ref{fig:2}, we group a wikiHow article and any video about the same task as a pair. Thanks to the inherent hierarchical structure of wikiHow articles, we can easily obtain sentences from different scales: \textbf{\textit{high-level}} summaries and \textbf{\textit{low-level}} details. As in Figure~\ref{fig:2}, ``\textit{Pour the lemon juice into the pitcher.}'' is a high-level sentence summary and ``\textit{You may add the pulp if .... along with the seeds.}'' is a low-level sentence detail of this summary. In this section, we first introduce the details of dataset construction, and then compare YouwikiHow to existing VG benchmarks.

\subsection{Dataset Construction}

\subsubsection{Training Set}

\noindent\textbf{Initial Visual Tasks.} Each wikiHow article describes a sequence of steps to instruct humans to perform a certain ``task'', and these tasks range from physical world interactions to abstract mental well-being improvement. In YouwikiHow, we follow~\cite{miech2019howto100m} and only focus on ``visual tasks". This gives us 25K tasks to begin with.

\noindent\textbf{Task-Related Videos.} We also follow~\cite{miech2019howto100m} and use the same preprocessing steps (\eg, remove videos with few views or too short durations) to obtain initial task-related videos for each task. To further control the quality and ensure sufficient training videos for each task, we restrict the videos to top 50 search results, and the number of training videos for each task to be at least 30. This step prunes the number of tasks from 25K to 2.3K.

\noindent\textbf{Sentence Quality Control.} Firstly, to avoid over-long articles, we filter out all the tasks with verbose sentences. Specifically, we set the max number of sentence summaries and details to 10 and 30, respectively. This filtering step decreases the task number to 1.4K. Meanwhile, since original wikiHow articles usually contain unimportant modifiers or quantifiers, we further conduct rule-based \emph{sentence simplification}~\cite{al2021automated} based on POS and dependency parse tags\footnote{For example, given the original sentence ``\textit{Then, mix in \textcolor{gray}{1 teaspoon (4.9 mL) of} vanilla extract, followed by \textcolor{gray}{1 teaspoon (2.6 grams) of} cinnamon.}'', sentence simplification can prune these unimportant modifier (\textcolor{gray}{gray} words) and obtain a new sentence: ``\textit{Then, mix in vanilla extract, followed by cinnamon.}''}.

\subsubsection{Test Set}

For the test set, we directly build on top of the existing CrossTask~\cite{zhukov2019cross} and reuse their manual temporal grounding annotations. Specifically, CrossTask is originally proposed for step segmentation, which consists of 18 primary wikiHow tasks. For each task, it collects corresponding YouTube videos and annotates the temporal grounding boundaries for each video corresponding to the predefined task-specific steps. Then, we manually link the step to the wikiHow articles\footnote{For example, we can easily link CrossTask steps ``brake on'' (\textit{Change a Tire}) or ``attach shelve'' (\textit{Build Simple Floating Shelves}) to the sentence ``\textit{Apply the parking brake and put car into `Park' position.}'' or ``\textit{Attach the shelf mount to the wall}'' in their corresponding wikiHow articles, respectively.} and propagate these annotations as the ground-truth for wikiHow sentences. We conduct the same sentence simplification steps on all the wikiHow articles in the test set, and remove the task with over-long articles\footnote{We remove three tasks: \textit{Make Kimichi Fried Rice}, \textit{Add Oil to Your Car}, and \textit{Make French Strawberry Cake}. All these tasks have over 60 sentences in their wikiHow articles.}. Unfortunately, when we perform manually linking between CrossTask steps and wikiHow articles, we found it is difficult to link these steps to low-level details and almost all steps are linked to high-level summaries. To this end, we further design different evaluation metrics for high-/low- level sentences to bypass these limitations. (Details are in Sec.~\ref{sec:5.1}.)

\subsection{Comparison with Existing VG Datasets}
We compare our collected YouwikiHow with other prevalent VG or step segmentation datasets in Table~\ref{tab:dataset_comparison}. In the training set, we have a total of 1,398 wikiHow tasks, and each task has an average of 33.88 videos. For each task, there are 6.01 high-level sentence summaries and 14.79 low-level sentence details. Compared to existing VG datasets, YouwikiHow has more training videos (47K vs. 21.8K), and much more query sentences for each video (20.8 vs. 7.7). More importantly, it supports multi-scale queries and the query sentences may not be grounded in the video. Compared to step segmentation datasets, it not only has much more queries for each video and more diverse training tasks (1,398 vs. 180), but also supports both open-vocabulary queries and multi-scale queries.






\begin{figure}[b]
\centering
  \vspace{-1.5em}
  \includegraphics[width=0.9\linewidth]{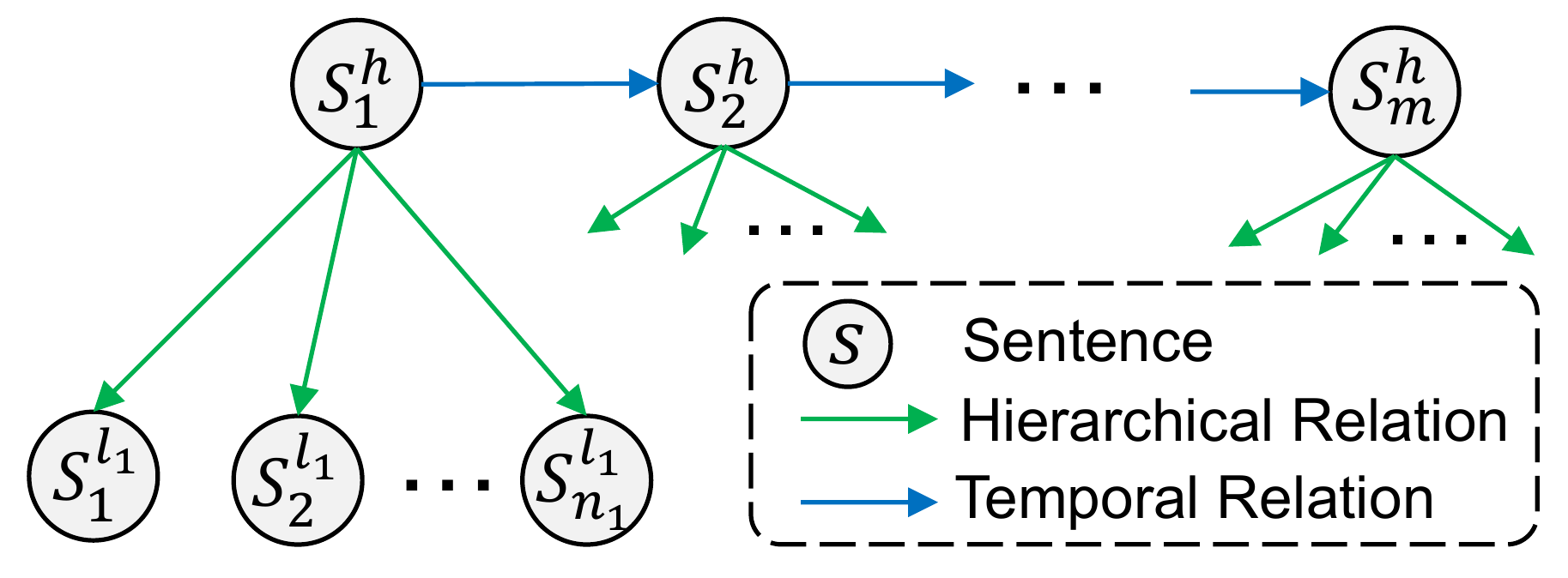}
  \vspace{-1em}
  \caption{Illustration of the multi-scale structures of $A$.}
  \label{fig:3}
\end{figure}

\begin{figure*}[t]
\centering
  \includegraphics[width=\linewidth]{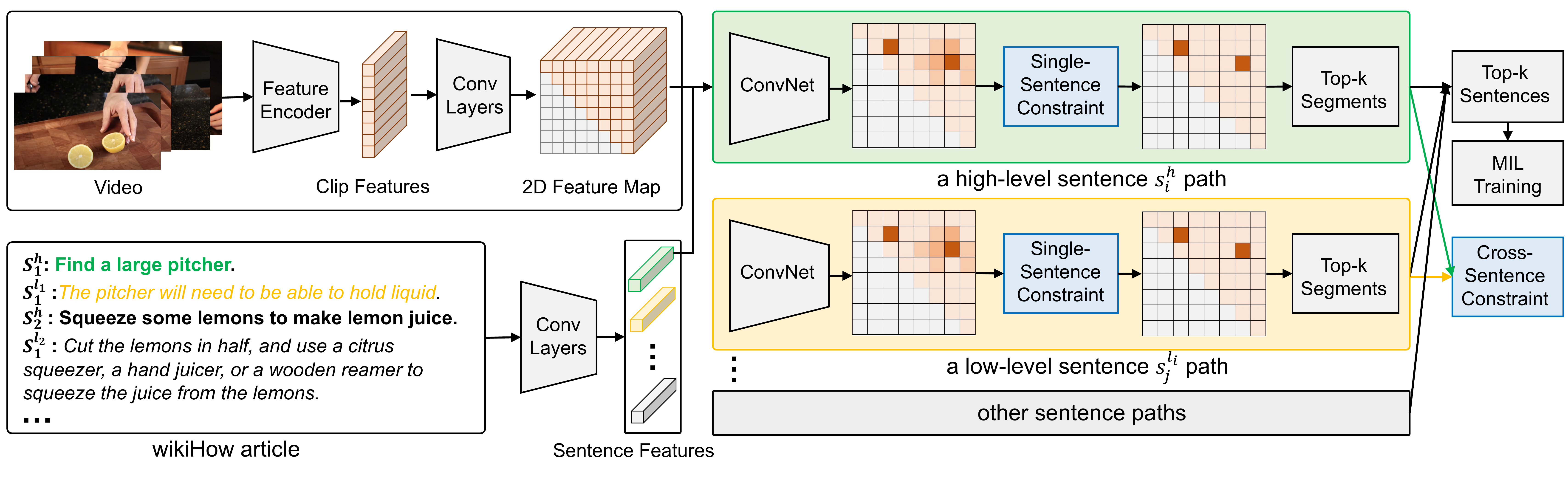}
  \vspace{-2em}
  \caption{The overview of the article grounding architecture with the proposed DualMIL.}
  \label{fig:4}
\end{figure*}

\section{Proposed Approach for WSAG}

\noindent\textbf{Problem Formulation.} WSAG is defined as follows: Given an untrimmed video $V$ and a relevant article $A$ with multi-scale sentences, WSAG needs to predict all possible temporal locations for all groundable sentences, \ie, one sentence may refer to either multiple segments or even none.

In this paper, we consider sentences at two scales. Specifically, as shown in Figure~\ref{fig:3}, article $A$ is organized as $ A = \{s^h_1, s^{l_1}_1, ..., s^{l_1}_{n_1}; s^h_2, ...; s^h_m, ..., s^{l_m}_{n_m} \}$, where $s^h_k$ is the $k$-th high-level summary, and $s^{l_k}_{i}$ is the $i$-th low-level details of $s^h_k$. There are $m$ high-level summaries in total, and each high-level summary $s^h_k$ has $n_k$ low-level details. To show more generalized abilities, in test stage, we assume that we do not know the scale prior of each sentence.

In this section, we first go through the architecture for grounding in Sec~\ref{sec:4.1}. Then, we detail each component of DualMIL in Sec~\ref{sec:4.2}.

\subsection{Basic Visual Grounding Architecture} \label{sec:4.1}

Since DualMIL is a model-agnostic training strategy, we follow a SOTA proposal-based model 2D-TAN~\cite{zhang2020learning} and use it as our baseline. As shown in Figure~\ref{fig:4}, it consists of three parts:

\noindent\textbf{Video Feature Encoding.} Given video $V$, we first use a pretrained video feature extractor to extract clip features, and sample the video features evenly to $N$ clips. Then, we utilize the 2D-map proposal strategy: All the segment proposals can be organized into a 2D temporal map $M$, and each element $m_{ij} \in M$ represents the candidate segment which starts from clip$_i$ and ends at clip$_j$. We extract each proposal feature by averaging all inside clip features, and then stack a few conv-layers to further encode the context. Finally, we obtain 2D feature map $\bm{F}^M \in \mathbb{R}^{N\times N\times d_v}$, and each element $\bm{F}^M_{ij}$ denotes the feature of segment proposal $m_{ij}$.

\noindent\textbf{Text Feature Encoding.} For each sentence $s_i = \{w^i_j\}$ in article $A$, we first use the GloVe embedding~\cite{pennington2014glove} to encode each word $w$, and then feed all word embeddings into a Bi-LSTM. The final hidden state of Bi-LSTM is taken as the feature of sentence, denoted as $\bm{F}^{S_i} \in \mathbb{R}^{d_s}$.

\noindent\textbf{Multimodal Matching.} After obtaining the video feature $\bm{F}^M$ and all sentence features $\{\bm{F}^{S_i}\}$, we then fuse these two features by Hadamard product:
\begin{equation}
    \bm{\tilde{F}}_{ij,k} = \bm{w}_s \bm{F}^{S_k} \odot \bm{w}_v \bm{F}^M_{ij},
\end{equation}
where $\bm{w}_s \in \mathbb{R}^{d_h \times d_s}$ and $\bm{w}_v \in \mathbb{R}^{d_h \times d_v}$ are two learnable MLPs, which map two modality features into a common space. Reorganizing the $\bm{\tilde{F}}_{ij,k}$ into the 2D map format, we can obtain $\bm{\tilde{F}}_k \in \mathbb{R}^{N\times N\times d_h}$, which denotes the fused feature between sentence $s_k$ and all segment proposals $M$.

Later, we adopt several conv-layers to obtain context-aware multimodal 2D feature maps. And these feature maps are fed into the classifier to predict all the matching score maps $\{\bm{P}^k\}$, where $\bm{P}^k \in \mathbb{R}^{N\times N}$ denotes the matching scores between all segment proposals $M$ and sentence $s_k$.

\subsection{DualMIL Training Objectives \& Inference} \label{sec:4.2}

\subsubsection{Two-level MIL Training Objective}
Since not all sentences in article $A$ are groundable to the given video $V$, we only select the top-$k_1$ sentences with the highest matching scores to represent the whole article. As for the matching score between each sentence and the video, we average the similarity scores among the top-$k_2$ proposals:
\begin{equation} \label{eq:2}
\begin{aligned}
    sim(V, s_k) &= avg\{ \text{top-}k_2 \max_{ij} \bm{P}^k_{ij} \},
\end{aligned}
\end{equation}
where $sim(V, s_k)$ denotes the similarity score between video $V$ and sentence $s_k$. Similarly, we use $sim(V, A)_i$ to denote the similarity score between the top-$i$ sentence in $A$ with video $V$ (\ie, $i \leq k_1$).

We train the whole model with the ranking loss. Specifically, we treat video $V$ and its same-task article $A$ as \emph{positive} pair ($V$, $A$). Then we randomly replace the video or article with other-task videos or articles to obtain \emph{negative} pairs, denoted as ($V^-$, $A$) and ($V$, $A^-$) respectively. Then, the two-level MIL loss is written as $\mathcal{L}_{\text{MIL}} = \sum_{i} \sum_{j} \mathcal{L}^{ij}_{\text{MIL}}$, and
\begin{equation} \label{eq:3}
\small
\begin{aligned}
\mathcal{L}^{ij}_{\text{MIL}} & = \max(0, \Delta - \text{sim}(V, A)_i + \text{sim}(V^-, A)_j)  \\
& +  \max(0, \Delta - \text{sim}(V, A)_i + \text{sim}(V, A^-)_j),
\end{aligned}
\end{equation}
where $\Delta$ is a predefined margin.

\subsubsection{Single-Sentence Constraint}

Since each sentence may be grounded in multiple segments, we need to predict the similarity scores between each query sentence and all segment proposals. To force WSAG models to make sparse predictions, we propose the single-sentence constraint to enhance the two-level MIL training. By ``sparse'', we mean that only a few proposals are selected as results for each groundable sentence.

Specifically, before selecting the top-$k_2$ segment proposals as in Eq.~\eqref{eq:2}, we conduct a sparse filtering step to suppress (or filter out) the proposals by two rules: 1) In local highly-overlapped neighbors, there are other proposals with higher video-sentence matching scores. 2) The matching score is much less than the proposal with the highest score.

From an implementation perspective, we can use a simple max-pooling layer with kernel size $K$ and a threshold $\delta$ to realize single-sentence constraint. Then, we can obtain a new filtered $\tilde{\bm{P}}^k$, and calculate a similar MIL loss with $\tilde{\bm{P}}^k$ following Eq.~\eqref{eq:2} and Eq.~\eqref{eq:3}. (Ablations on $K$ and $\delta$ are in Sec.~\ref{sec:5}).

\noindent\textbf{Highlights.} Compared to the existing constraint strategy by selecting the proposal with the highest score as extra pseudo GT~\cite{wang2021weakly}, our solution avoids selecting unstable pseudo GT (\ie, more robust), and it is more suitable for the setting of any number of GT segments for each query.

\subsubsection{Cross-Sentence Constraint} \label{sec:4.2.3}

To force WSAG models to perceive multi-scale queries, we propose the cross-sentence constraint. Specifically, we assume that high-level sentences should be more like to grounded than its low-level sentences for highly matched proposals. The reason is that today's multimodal coreference relations between query sentence and GT video segment discussed in grounding works contain both ``\emph{identical}'' and ``\emph{hierarchical}'' relations. Let's take two extreme cases as examples: 1) If the low-level sentence is identical to the video segment proposal, then its high-level sentence is also coreference to the proposal (hierarchical relation). 2) If the high-level sentence is identical to the proposal, then its low-level sentence is only partially coreference to the proposal. Thus, we propose the cross-sentence constraint by limiting the proposal matching scores between a high-level and low-level sentence pair.

Obviously, if the proposal itself is unrelated to the low-level sentence, this constraint is meaningless. Thus, we use the low-level sentence matching score as the loss weight, and the contraint loss is:
\begin{equation}
\small
    \mathcal{L}_{\text{CS}} = \sum_{h=1}^m \sum_{k=1}^{n_h} \sum_{ij}\max(0, \alpha - \bm{P}^{h}_{ij} + \bm{P}^{l_h, k}_{ij}) \cdot \bm{P}^{l_h, k}_{ij},
\end{equation}
where $\bm{P}^{h}_{ij}$ is the matching score between $h$-th high-level sentence and proposal $m_{ij}$, and $\bm{P}^{l_h, k}_{ij}$ is the matching score between $k$-th low-level sentence of $s^h_h$ and proposal $m_{ij}$. $\alpha$ is a predefined margin, and the impact of $\alpha$ is discussed in Table~\ref{tab:CS-Const}.

\noindent\textbf{Highlights.} Since multimodal hierarchical relation is always difficult to predict, the main effect of the cross-sentence constraint is to avoid the case: a low-level sentence has a high matching score with the proposal while its high-level summary is not.

\subsubsection{Inference} 
In the test stage, given a video and a relevant article, we first predict the matching scores between each sentence and all proposals, and then we conduct non-maximum suppression (NMS) to filter out the proposals with highly overlaps but smaller scores. Then, we can simply combine all predictions from different sentences based on their matching scores.

To further consider the semantic relations between sentences at test stage, we use a \textbf{Structure-NMS}, inspired by Soft-NMS~\cite{bodla2017soft}, to suppress the segments which violate structure constraints. More details are left in the appendix.

\section{Experiments} \label{sec:5}

\subsection{Experimental Settings} \label{sec:5.1}

\noindent\textbf{Metrics.}  We used \emph{Recall@K} (\textbf{R@K}) over different IoU thresholds (0.1/0.3/0.5) to evaluate each video-article pair. Specifically, we ranked all segment-sentence pairs based on their matching scores, and calculated the recalls of all GT annotations within top-K predictions\footnote{We used Recall as metrics for two reasons: 1) Following existing VG works, they also use Recall@K as the main metric. 2) Due to our GT propagation rules for the test set, we may miss some GT annotations, \ie, (cf. Sec. Limitations).  }. A prediction is hit if its IoU with GT is larger than the threshold. Since we only have GT annotations for high-level summaries, we also proposed \emph{Recall@K meet Constraint} (\textbf{RC@K}) as a supplementary metric for low-level sentences. Since we assume the temporal grounding results of low-level sentences should be inside its high-level manual annotations, we calculated the percentages of low-level sentence predictions that meet the constraint. Note that RC@K is not strictly accurate.

\noindent\textbf{Implementation Details.}
Given a reference video $V$, we used a pretrained S3D extractor~\cite{miech2020end} to extract initial clip features. The number of initial clips was set to 256. For text sentences, following prior VG works, we truncated or padded each sentence to a maximum length of 25 words. In the training stage, to save GPU memory, we randomly sample 20 sentences if the articles have more than 20 sentences. All the dimensions of the hidden features were set to 512. In the multimodal matching, we used a three-layer convolutional network to encode context. Its kernel size and strides were set to 3 and 1, respectively. We trained the whole network with Adam optimizer for 100 epochs. The initial learning rate was set to 0.0001, and the batch size was set to 32. The loss weights of two-level MIL loss (for both models with and without single-sentence constraint) and cross-sentence constraint loss were set to 1.0 and 0.1, respectively. The predefined margin $\Delta$ for MIL training was set to 0.3. For the model with cross-sentence constraint, to ensure the predicted low-level sentence matching scores are reliable, we first train the model with MIL loss solely at a warm-up stage, and then add cross-sentence constraint loss for further training.



\addtolength{\tabcolsep}{-3pt}
\begin{table}[t]
    \centering
    \scalebox{0.90}{
        \begin{tabular}{|l|c c c | c c c|}
        \hline
            \multirow{2}{*}{Model} & \multicolumn{3}{c|}{\textbf{R@50 (IoU)}} & \multicolumn{3}{c|}{\textbf{R@100 (IoU)}}  \\
             & 0.1 & 0.3 & 0.5 & 0.1 & 0.3 & 0.5  \\
        \hline
            Baseline & 26.60 & 14.98 & 6.48 & 44.05 & 24.81 & 10.82 \\
        \hline
            \multicolumn{7}{|l|}{Baseline w/ \emph{Single-Sentence Constraint}}  \\
        \hline
            $K$=7, $\delta$=0.9 & 27.86 & 16.00 & 7.00 & 41.76 & 23.95 & 10.63 \\
            $K$=7, $\delta$=0.7 & 30.02 & \textbf{17.38} & \textbf{7.70} & 43.50 & 25.50 & 11.68 \\
            $K$=7, $\delta$=0.5 & \textbf{30.37} & 17.30 & 7.51 & \textbf{47.57} & \textbf{27.21} & \textbf{11.97} \\
            $K$=5, $\delta$=0.5 & 27.36 & 15.15 & 6.67 & 45.24 & 25.62 & 11.37 \\
            $K$=3, $\delta$=0.5 & 26.25 & 14.66 & 6.66 & 45.99 & 25.95 & 11.49 \\
        \hline
        \end{tabular}
    }
    \vspace{-0.5em}
    \caption{Ablations (\%) on single-sentence constraint.}
    \vspace{-1em}
    \label{tab:SS-Const}
\end{table}
\addtolength{\tabcolsep}{3pt}

\addtolength{\tabcolsep}{-4pt}
\begin{table}[t]
    \centering
    \scalebox{0.95}{
        \begin{tabular}{|l|c c c | c c c|}
        \hline
            \multirow{2}{*}{Model} & \multicolumn{3}{c|}{\textbf{R@50 (IoU)}} & \multicolumn{3}{c|}{\textbf{R@100 (IoU)}}  \\
             & 0.1 & 0.3 & 0.5 & 0.1 & 0.3 & 0.5  \\
        \hline
            Baseline & 26.60 & 14.98 & 6.48 & 44.05 & 24.81 & 10.82 \\
        \hline
            \multicolumn{7}{|l|}{Baseline w/ \emph{Cross-Sentence Constraint}}  \\
        \hline
            $\alpha$ = 0.1 & 33.79 & 18.98 & 8.21 & 52.60 & 29.37 & 12.83 \\
            $\alpha$ = 0.0 & \textbf{34.73} & \textbf{19.23} & \textbf{8.46} & \textbf{55.07} & \textbf{30.66} & \textbf{13.46} \\
            $\alpha$ = -0.1 & 32.63 & 18.38 & 8.05 & 52.28 & 29.42 & 12.93 \\
        \hline
        \end{tabular}
    }
    \vspace{-0.5em}
    \caption{Ablations (\%) on cross-sentence constraint.}
    \vspace{-1em}
    \label{tab:CS-Const}
\end{table}
\addtolength{\tabcolsep}{4pt}

\subsection{Ablation Studies}

We run a number of ablations to analyze the impact of different hyperparameters of each component, and the effectiveness of each component.

\noindent\textbf{Ablation on Single-Sentence Constraint.} The impacts of two hyperparameters in the single-sentence constraint (\ie, kernel sizes $K$ and thresholds $\delta$) are reported in Table~\ref{tab:SS-Const}. From the results, we can observe that: 1) For most hyperparameter settings, the single-sentence constraint can consistently improve models' performance. 2) The Model with setting $K=7$ and $\delta=0.5$ achieves the best results.

\noindent\textbf{Ablation on Cross-Sentence Constraint.} The impact of different margin $\alpha$ in the cross-sentence constraint are reported in Table~\ref{tab:CS-Const}. From results, we can observe that the performance gains are robust to different $\alpha$, and the model with $\alpha=0$ achieves the best performance. It is worth noting that a negative $\alpha$ (relaxed constraint) is still effective, which proves the claimed main effects in Sec.~\ref{sec:4.2.3}.

\addtolength{\tabcolsep}{-3.5pt}
\begin{table}[t]
    \centering
    \scalebox{0.90}{
        \begin{tabular}{|c c c | c c c | c c c |}
        \hline
            \multicolumn{3}{|c|}{Strategies} & \multicolumn{3}{c|}{\textbf{R@50 (IoU)}} & \multicolumn{3}{c|}{\textbf{R@100 (IoU)}}   \\
             SS & CS & NMS & 0.1 & 0.3 & 0.5 & 0.1 & 0.3 & 0.5 \\
        \hline
            \XSolidBrush & \XSolidBrush & \XSolidBrush & 26.60 & 14.98 & 6.48 & 44.05 & 24.81 & 10.82  \\
            \XSolidBrush & \XSolidBrush & \Checkmark & 26.92 & 15.22 & 6.63 & 44.14 & 24.87 & 10.84  \\
            \Checkmark & \XSolidBrush & \XSolidBrush & 30.37 & 17.30 & 7.51 & 47.57 & 27.21 & 11.97  \\
            \XSolidBrush & \Checkmark & \XSolidBrush & 34.73 & 19.23 & 8.46 & \textbf{55.07} & 30.66 & 13.46  \\
            \Checkmark & \Checkmark & \Checkmark & \textbf{40.11} & \textbf{23.08} & \textbf{10.07} & 54.32 & \textbf{31.30} & \textbf{13.97} \\
        \hline
        \end{tabular}
    }
    \vspace{-0.5em}
    \caption{Ablations (\%) on the effectiveness of each part, where ``SS'', ``CS'', and ``NMS'' denote single-/cross- sentence constraint and structure-NMS, respectively.}
    \label{tab:ablations}
\end{table}
\addtolength{\tabcolsep}{3.5pt}

\begin{figure*}[htbp]
    \begin{minipage}[c]{0.28\linewidth}
        \centering
        \includegraphics[width=0.9\linewidth]{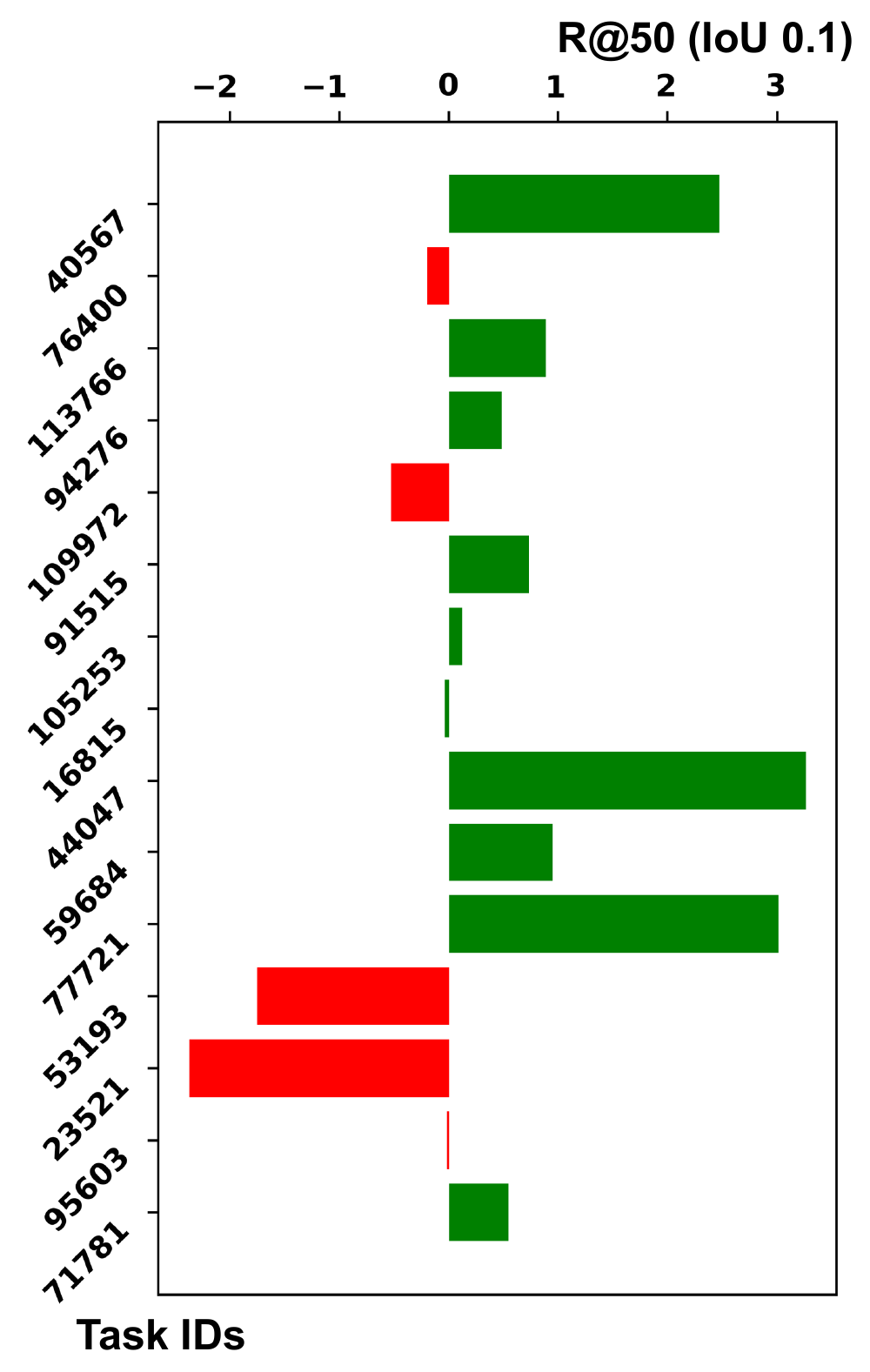}
        \vspace{-0.8em}
        \caption{Performance gains (\%) between models w/ \& w/o structure-NMS. Task ids are ranked by the agreement between the order of groundable sentences and GT segments (cf. Appendix).}
        \vspace{-0.5em}
        \label{fig:5}
        \addtolength{\tabcolsep}{4pt}
    \end{minipage} \hfill
    \begin{minipage}[c]{0.7\linewidth}
        \centering
        \includegraphics[width=0.98\linewidth]{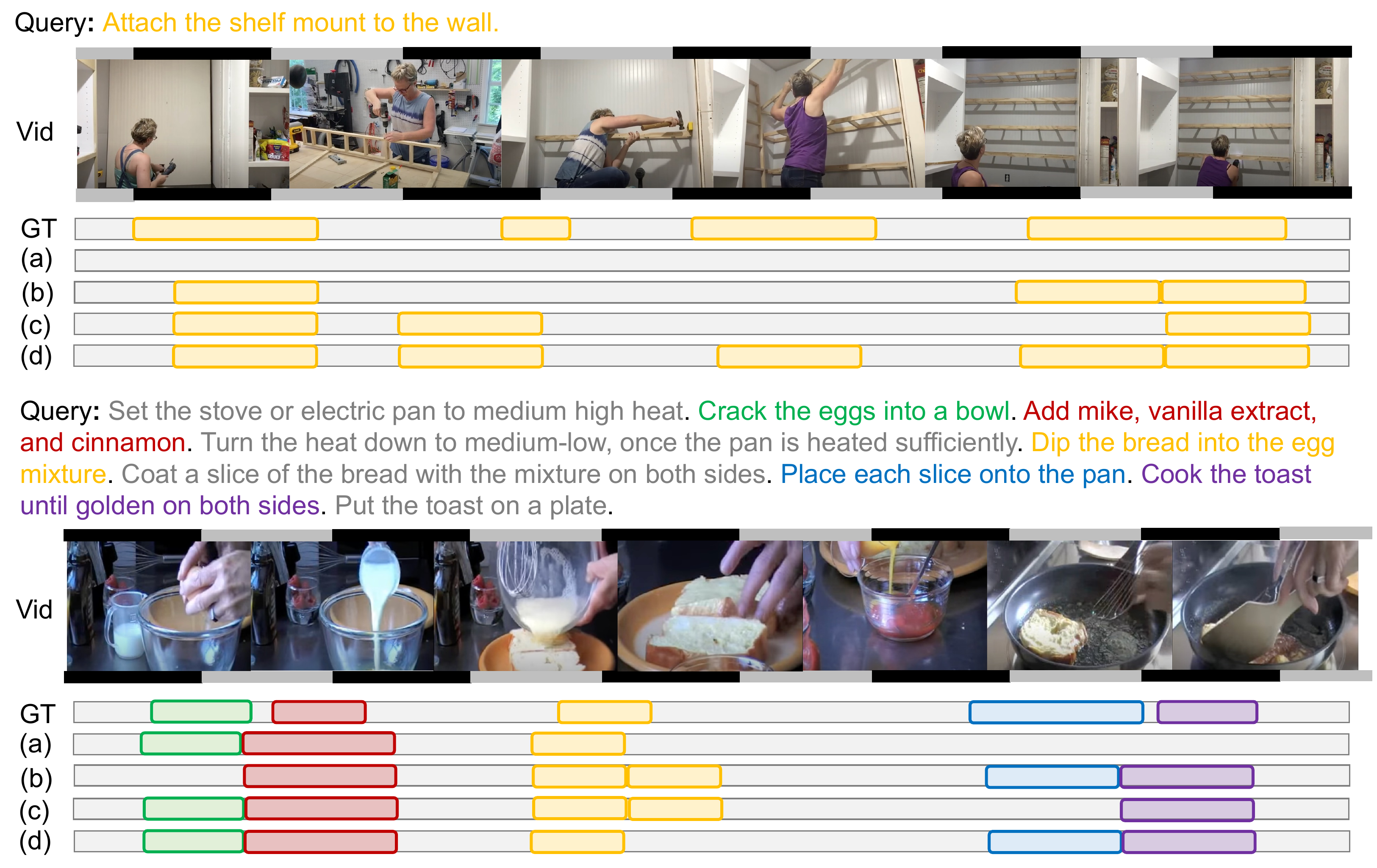}
        \vspace{-1em}
        \caption{\textbf{Upper}: An example of a query with multiple GT segments. \textbf{Below}: Given a video and an article (only high-level sentences), GT segments of all groundable sentences are shown (with corresponding colors). (a) - (d) denotes baseline, baseline w/ single-sentence const., baseline w/ cross-sentence const., and full model, respectively. Top-50 predictions overlapped with GT are shown.}
        \vspace{-0.5em}
        \label{fig:6}
    \end{minipage}
\end{figure*}

\addtolength{\tabcolsep}{-2.5pt}
\begin{table*}[t]
    \centering
    \scalebox{0.90}{
        \begin{tabular}{|c |l | c | c c c | c c c | c c c | c c c }
        \hline
            & \multirow{2}{*}{Model} & \multirow{2}{*}{MM Pretrain} & \multicolumn{3}{c|}{\textbf{R@50} (IoU)} & \multicolumn{3}{c|}{\textbf{R@100} (IoU)} & \multicolumn{3}{c|}{\textcolor{mygray}{\textbf{RC@50} (IoU)}}  \\
             & & & 0.1 & 0.3 & 0.5 & 0.1 & 0.3 & 0.5 & \textcolor{mygray}{0.1} & \textcolor{mygray}{0.3} & \textcolor{mygray}{0.5}  \\
        \hline
            \multirow{3}{*}{Type1} & RandomGuess$^*$ &  & 19.55 & 5.22 & 1.67 & 33.05 & 10.46 & 3.88 & \textcolor{mygray}{7.23} & \textcolor{mygray}{1.87} & \textcolor{mygray}{0.66} \\
            & MIL-XE (WSTAN$_{\text{base}}$) &  & 27.22 & 15.98 & 7.42 & 36.52 & 20.97 & 9.92 & \textcolor{mygray}{13.75} & \textcolor{mygray}{10.59} & \textcolor{mygray}{9.25} \\
            & WSTAN$^\dagger$ &  & 16.41 & 2.36 & 0.49 & 16.51 & 2.36 & 0.49 & \textcolor{mygray}{6.96} & \textcolor{mygray}{0.82} & \textcolor{mygray}{0.19} \\
        \cline{1-2}\cdashline{4-12}
            \multirow{2}{*}{Type2} & MIL-NCE-max & \Checkmark & 33.48 & 12.01 & 4.89 & 39.71 & 14.30 & 5.87 & \textcolor{mygray}{11.85} & \textcolor{mygray}{3.12} & \textcolor{mygray}{1.06} \\
            & MIL-NCE-avg & \Checkmark & \textbf{\textcolor{blue}{42.86}} & \textbf{\textcolor{blue}{24.26}} & \textbf{\textcolor{blue}{12.88}} & \textbf{\textcolor{blue}{56.77}} & \textbf{\textcolor{blue}{32.04}} & \textbf{\textcolor{blue}{16.98}} & \textcolor{mygray}{16.40} & \textcolor{mygray}{7.71} & \textcolor{mygray}{3.87} \\
        \cline{1-2}\cdashline{4-12}
            \multirow{3}{*}{Type3} & MIL-NCE+WSTAN$_{\text{base}}$ a & \Checkmark & 27.84 & 12.10 & 5.65 & 41.75 & 18.59 & 8.94 & \textcolor{mygray}{10.98} & \textcolor{mygray}{4.01} & \textcolor{mygray}{1.52} \\
            & MIL-NCE+WSTAN$_{\text{base}}$ b & \Checkmark & 33.94 & 15.01 & 7.01 & 47.56 & 21.43 & 10.50 & \textcolor{mygray}{9.23} & \textcolor{mygray}{3.68} & \textcolor{mygray}{1.78} \\
            & MIL-NCE+WSTAN$_{\text{base}}$ c & \Checkmark & 32.30 & 18.03 & 8.61 & 50.30 & 28.81 & \textcolor{red}{14.04} & \textcolor{mygray}{13.44} & \textcolor{mygray}{10.49} & \textcolor{mygray}{9.09} \\
        \hline
            & \textbf{DualMIL (Ours)} &  & \textcolor{red}{40.11} & \textcolor{red}{23.08} & \textcolor{red}{10.07} & \textcolor{red}{54.32} & \textcolor{red}{31.30} & 13.97 & \textcolor{mygray}{10.71} & \textcolor{mygray}{8.09} & \textcolor{mygray}{6.96} \\
        \hline
        \end{tabular}
    }
    \vspace{-0.5em}
    \caption{Performance (\%) comparison with SOTA baselines. All listed methods use the same proposal settings. ``MM Pretrain'' denotes these models use large-scale multimodal pretraining features. $^*$ results are averaged by five different random seeds. Model a/b/c in Type3 denotes model with different thresholds. $^\dagger$ denotes reimplementation results using official codes. The \textbf{\textcolor{blue}{best}} and \textcolor{red}{second best} results are denotes with corresponding formats.}
    \label{tab:SOTA}
\end{table*}
\addtolength{\tabcolsep}{2.5pt}

\noindent\textbf{Ablation on Structure-NMS.} The results of the models with and without structure-NMS are illustrated in Figure~\ref{fig:5}. From the results, we can observe that structure-NMS can significantly improve the performance of tasks with high agreements (\eg, \textit{Change a Tire}, or \textit{Grill Steak}). In contrast, it may hurt the performance of tasks with low agreements.

\noindent\textbf{Effectiveness of Each Strategy.} The ablation studies on each strategy are reported in Table~\ref{tab:ablations}. From Table~\ref{tab:ablations}, we have the following observations: 1) Compared to the baseline, each strategy can consistently improve performance on both R@50 and R@100 metrics. 2) The full model achieves the best R@50 and R@100 over different IoUs.

\subsection{Comparisons with State-of-the-Art}

\noindent\textbf{Baselines.} We compared our proposed DualMIL with a set of state-of-the-art baselines. Specifically, we investigated three types of baselines:

\noindent\textbf{\textit{Type1}}: State-of-the-art WSVG models. We compared with \textbf{WSTAN}~\cite{wang2021weakly}, which builds on top of a cross-entropy (XE) based MIL backbone. For completeness, we also reported results of the WSTAN backbone (dubbed \textbf{MIL-XE}), and a random guess (\textbf{RandomGuess}) baseline.

\noindent\textbf{\textit{Type2}}: Pretrained multimodal video-text retrieval models (\eg, \textbf{MIL-NCE}~\cite{miech2020end}). We show the \emph{zero-shot} results of two variants by max-pooling or average-pooling the clip features inside the boundaries of video segment proposals.

\noindent\textbf{\textit{Type3}}: Two-stage model. Since today's WSVG models assume all the sentences can be grounded to the video, a straightforward two-stage solution is: Using pretrained video-text retrieval models to select all groundable sentences first, and then training a WSVG model with selected sentences. Obviously, we need to manually set a threshold to filter out sentences at the first stage, we reported results of three variants with different thresholds.

\noindent\textbf{Results.} All results are reported in Table~\ref{tab:SOTA}. From Table~\ref{tab:SOTA}, we have the following observations: 1) For Type1 methods, the simple baseline MIL-XE can achieve good performance. However, the SOTA model WSTAN with other more advanced designs only performs similarly with RandomGuess, which proves existing SOTA WSVG models fail to work in these more realistic settings. 2) For Type2 methods, the performance gaps between different pooling operations are large. Although these large-scale pretrained models can achieve exemplary zero-shot performance, they are not robust enough and heavily rely on different heuristic rules. 3) For Type3 methods, the model with different thresholds also behavior differently, \ie, these two-stage methods are not robust either. 4) In contrast, our proposed DudalMIL can achieve satisfactory performance with relatively consistent gains.

\subsection{Visualizations}

We illustrated two examples in Figure~\ref{fig:6}. For the first example, we only show the grounding results of one query sentence (from article ``\textit{Build Simple Floating Shelves}'') with multiple ground-truth segments. For the second example, we show the grounding results of all high-level sentences of the article (``\textit{Make French Toast}''). From Figure~\ref{fig:6}, we observe that: Both the proposed single-sentence constraint and cross-sentence constraint can help to ground some missing segments in top-K predictions. Meanwhile, both constraints are complementary, \ie, the full model achieves the best results.

\section{Conclusions}
In this paper, we discussed the weaknesses of default assumptions in existing video grounding work, and proposed a more challenging task: weakly-supervised article grounding (WSAG). To facilitate the research in this direction, we collected the first WSAG dataset YouwikiHow. Further, we proposed DualMIL for WSAG, including a two-level MIL loss and a single-/cross-sentence constraint loss. This work paves the way for a number of exciting future works: 1) designing more reasonable backbones for multiple sentence inputs by considering their semantic relations; 2) extending to more general domains beyond instructional articles.

\section*{Acknowledgments} \vspace{-0.5em}
We thank the anonymous reviewers' helpful suggestions. This research is based upon work supported by U.S. DARPA KAIROS Program No. FA8750-19-2-1004. The views and conclusions contained herein are those of the authors and should not be interpreted as necessarily representing the official policies, either expressed or implied, of DARPA, or the U.S. Government. The U.S. Government is authorized to reproduce and distribute reprints for governmental purposes notwithstanding any copyright annotation therein.

\section*{Limitations} \vspace{-0.5em}
The main limitations of this work are about the collected dataset \emph{YouwikiHow}. Specifically, we can discuss them from the two following aspects:

\textbf{Dataset Creation.} Since we focus on WSAG, the manner of creating the training set of YouwikiHow is acceptable. However, to save the manual annotations for the test set, we only propagate the annotations from the existing CrossTask~\cite{zhukov2019cross} dataset. Although this solution is much cheaper, it introduces two types of potential errors in the ``ground-truth'' annotations for evaluation: 1) When manually mapping the ``step'' in CrossTask to the ``sentence'' in the wikiHow article, we found it not always be one-to-one perfect mapping. In a few cases, multiple sentences may refer to a single step or multiple steps may refer to a single sentence. Thus, the original ground-truth annotations for each CrossTask step may not be exactly accurate for its mapped sentence regarding the same video. 2) Since each wikiHow article has much more sentence queries than original step queries in CrossTask, many wikiHow sentences cannot be mapped to these predefined steps, \ie, these wikiHow sentences will not have any ``ground-truth'' annotations. However, these sentences may be groundable in some specific videos.
    
\textbf{Domain Coverage.} Since we obtain the explicit multi-scale sentences from the inherent hierarchical structures of wikiHow articles, these wikiHow articles are mainly about instructional articles. Thus, the main domain of our YouwikiHow dataset focuses on instructional articles/videos, \ie, the model trained in our dataset may suffer from performance drops when they are applied to other domain daily multimodal assets. 

For the first limitation, we mitigate its impact by using more relaxed metrics: R@K or RC@K. Of course, the most accurate solution is checking all the annotations between any video-article pairs.



\section*{Ethics Statement}\vspace{-0.5em}
The proposed dataset and method aim to improve the performance of temporal grounding models in more realistic settings. Advancements in visual grounding help the deployment of visual grounding (or article grounding) models in our daily applications. Since we mainly focus on the two unrealistic assumptions in existing grounding models, our work does not introduce new ethical concerns. The only potential ethical concern is that any language-query based applications run the risk of using biased or offensive words (or descriptions) --- video grounding is no exception. In the future, we can try to incorporate a preprocessing step to avoid or correct biased or offensive content.



\bibliography{emnlp2022}
\bibliographystyle{acl_natbib}

\section*{Appendix}
\appendix

The appendix is organized as follows:
\begin{itemize}
    \item More details about the structure-NMS are in Sec.~\ref{sec:a}.
    
    \item More experimental details are in Sec.~\ref{sec:b}.
    
    \item More ablation studies are in Sec.~\ref{sec:c}.
    
    \item The statistics about the agreement of GT temporal orders are in Sec.~\ref{sec:d}.
\end{itemize}

\section{More Details about Structure-NMS} \label{sec:a}
Given all detected segments for each groundable sentence in the article, we hope these segments themselves also meet the same semantic relations as their query sentences (temporal or hierarchical relations). Since we assume that we do not know the scale prior of each sentence at the test stage, currently we only consider the temporal relations.

More specifically, given two query sentences $s_i$ and $s_j$. If $s_i$ appears earlier than $s_j$ in the corresponding article, we hope the grounding segments for $s_i$ should be earlier than $s_j$ too. Following Soft-NMS~\cite{bodla2017soft}, we also multiple a coefficient to decrease the matching score of the proposals which violate this temporal constraint, and the coefficient is proportional to their IoU. Let's take a concrete example. If the predicted segments for $s_i$ and $s_j$ are $[l^i_s, l^i_e]$ and $[l^j_s, l^j_e]$, and their matching scores are $p^i$ and $p^j$ ($p^i < p^j$). After selecting the segment $[l^j_s, l^j_e]$ into top-K predictions, we then slightly decrease the matching score $p^i$ by:
\begin{equation}
\small
\begin{aligned}
    IoU_{bad} &= \frac{\max(l^j_s- l^i_s, 0) + \max(l^i_e - l^j_e, 0)}{\max(l^i_e, l^j_e) - \min(l^i_s, l^j_s)}, \\
    p^i_{new} &= p^i * \exp(-(IoU_{bad})**2 / \text{const}), \\
\end{aligned}
\end{equation}
where const is a constant number.

\section{More Experimental Details} \label{sec:b}

\noindent\textbf{More Details about RC@K.} Since we hope the grounding segment of the low-level sentence is inside that of their high-level summary, we calculate RC@K the same way as plain recall with only one exception: if the low-level prediction is totally inside their high-level ground-truth annotations, the prediction is regarded as hit regardless of the IoU.

\addtolength{\tabcolsep}{-3pt}
\begin{table}[t]
    \centering
    \scalebox{0.90}{
        \begin{tabular}{|c|l|c c c | c c c|}
        \hline
            & \multirow{2}{*}{Setting} & \multicolumn{3}{c|}{\textbf{R@50}} & \multicolumn{3}{c|}{\textbf{R@100}}  \\
            & & 0.1 & 0.3 & 0.5 & 0.1 & 0.3 & 0.5  \\
        \hline
            \parbox[t]{3mm}{\multirow{4}{*}{\rotatebox[origin=c]{90}{\textbf{Prediction}}}} & N=8 & 26.57 & 10.69 & 4.13 & 43.55 & 17.39 & 6.73  \\
            & N=12 & 26.01 & 13.46 & 4.64 & \textbf{44.15} & 21.61 & 9.01  \\
            & N=16 & \textbf{26.60} & \textbf{14.98} & \textbf{6.48} & 44.05 & 24.81 & 10.82 \\
            & N=24 & 24.69 & 14.81 & 6.24 & 39.79 & \textbf{25.25} & \textbf{11.29} \\
        \hline
            \parbox[t]{3mm}{\multirow{4}{*}{\rotatebox[origin=c]{90}{\textbf{GT}}}} & N=8 & 64.14 & 26.88 & 11.57 & 66.82 & 28.10 & 12.29 \\
            & N=12 & 68.02 & 35.31 & 15.77 & 74.06 & 38.24 & 17.52 \\
            & N=16 & \textbf{71.79} & \textbf{42.47} & \textbf{19.90} & \textbf{80.86} & 47.88 & 23.00 \\
            & N=24 & 62.45 & 41.07 & 19.29 & 76.05 & \textbf{50.43} & \textbf{24.03} \\
        \hline
        \end{tabular}
    }
    \vspace{-0.5em}
    \caption{Ablations on different proposal settings. ``GT'' denotes the results with only groundable sentences.}
    \label{tab:proposal}
\end{table}
\addtolength{\tabcolsep}{3pt}

\begin{table}[t]
    \centering
    \begin{tabular}{l|l|c}
        \hline
            \textbf{ID} & \textbf{Task Name} & \textbf{Agree.} \\
        \hline
            40567 & \small{Change a Tire} & 96.94\% \\
            76400 & \small{Make French Toast} & 93.11\% \\
            113766 & \small{Grill Steak} & 89.33\% \\
            94276 & \small{Make Meringue} & 86.71\% \\
            109972 & \small{Make Banana Ice Cream} & 86.89\% \\
            91515 & \small{Make Pancake} & 85.49\% \\
            105253 & \small{Make Bread and Butter Pickles} & 83.61\% \\
            16815 & \small{Jack Up a Car} & 78.01\% \\
            44047 & \small{Make Lemonade} & 77.67\% \\
            59684 & \small{Build Simple Floating Shelves} & 75.39\% \\
            77721 & \small{Make Irish Coffee} & 71.21\% \\
            53193 & \small{Make a Latte} & 69.64\% \\
            23521 & \small{Make Jello Shots} & 68.61\% \\
            95603 & \small{Make Kerala Fish Curry} & 62.75\% \\
            71781 & \small{Make Taco Salad} & 54.39\% \\
        \hline
    \end{tabular}
    \vspace{-0.5em}
    \caption{The statistics about the agreement between the order of ground-truth query sentence and the order of their corresponding ground-truth segments.}
    \label{tab:task_agreement}
\end{table}

\section{More Ablation Studies} \label{sec:c}

\noindent\textbf{Impact of Proposal Settings.} For proposal-based VG methods, a notorious weakness is that their performance is heavily affected by different proposal settings. To this end, we explored the impact of different proposal settings in our baseline framework, and the results are reported in Table~\ref{tab:proposal}. From Table~\ref{tab:proposal}, we can observe that the model achieves the best performance in most metrics when $N$ is 16. The performance gap between the ``prediction'' and ``GT'' settings also shows that the main bottleneck for current article grounding models is detecting groundable sentences for the video-article pair.

\section{Statistics about the Agreement of GT Temporal Orders} \label{sec:d}

The agreement between the order of all groundable sentences and the order of their corresponding ground-truth segments of the test set are reported in Table~\ref{tab:task_agreement}.


\end{document}